\documentclass{article}

\usepackage{microtype}
\usepackage{graphicx}
\usepackage{subcaption}
\usepackage{booktabs} 

\usepackage[dvipsnames]{xcolor}         
\definecolor{navy}{rgb}{0.0, 0.0, 0.5}
\definecolor{ourgreen}{RGB}{118, 185, 0} 
\usepackage[backref]{hyperref}

\usepackage[preprint]{icml2026}

\usepackage{amsmath}
\usepackage{amssymb}
\usepackage{mathtools}
\usepackage{amsthm}
\usepackage{amsfonts}
\usepackage{bm}
\usepackage{braket}
\usepackage{nicefrac}

\usepackage[capitalize,noabbrev]{cleveref}

\theoremstyle{plain}

\theoremstyle{definition}

\theoremstyle{remark}

\usepackage[textsize=tiny]{todonotes}

\newcommand{\mole}{M\=oLe}
\newcommand{\molelambda}{M\=oLe-\texorpdfstring{$\Lambda$}{Lambda}}
\newcommand{\ccsd}{CCSD}

\icmltitlerunning{M\=oLe-$\mathbf{\Lambda}$: Learning the Coupled-Cluster Response State}

\begin{document}

\title{}
\twocolumn[
  \icmltitle{M\=oLe-$\mathbf{\Lambda}$: Learning the Coupled-Cluster Response State for Energies, Gradients, and Properties}



  \icmlsetsymbol{equal}{*}

  \begin{icmlauthorlist}
      \icmlauthor{Andreas Burger}{uoft,vector,nvidia}
      \icmlauthor{Luca Thiede}{uoft,vector,nvidia}
      \icmlauthor{Abdulrahman Aldossary}{nvidia}
      \icmlauthor{Jorge Arturo Campos-Gonzalez-Angulo}{uoft,vector}
      \icmlauthor{Alexander Zook}{nvidia}
      \icmlauthor{Jérôme Florian Gonthier}{nvidia}
      \icmlauthor{Alán Aspuru-Guzik}{uoft,vector,accel,cifar}
    \end{icmlauthorlist}

    \icmlaffiliation{uoft}{University of Toronto, 
    Canada}
    \icmlaffiliation{vector}{Vector Institute for Artificial Intelligence, 
    Toronto, Canada}
    \icmlaffiliation{nvidia}{NVIDIA, 
    Toronto, Canada}
    \icmlaffiliation{accel}{Acceleration Consortium, 
    Toronto, Canada}
    \icmlaffiliation{cifar}{
    Senior Fellow, Canadian Institute for Advanced Research (CIFAR), 
    Toronto, Canada
    }
    
    \icmlcorrespondingauthor{Andreas Burger}{me@andreas-burger.com}

  \icmlkeywords{Machine Learning, ICML}

  \vskip 0.3in
]



\printAffiliationsAndNotice{}  

\begin{abstract}
Coupled-cluster (CC) theory is often considered the gold standard of quantum chemistry, but its high computational cost limits routine access to accurate energies, forces and response properties.
While the right-hand $T$-amplitudes determine the correlated wavefunction, many practically important observables additionally require the left-hand $\Lambda$-amplitudes.
We introduce \molelambda, an extension of Molecular Orbital Learning (\mole) that predicts the full ground-state coupled-cluster singles and doubles (CCSD) response state by jointly learning right-hand amplitudes $(T_1,T_2)$ and left-hand amplitudes $(\Lambda_1,\Lambda_2)$ from localized Hartree--Fock molecular orbitals. 
Architecturally, \molelambda{} extends \mole{} with $\Lambda_1$ and $\Lambda_2$ readouts that mirror the symmetry constraints of the $T_1$ and $T_2$ heads, while preserving the original equivariant orbital encoder, odd sign-equivariant decoding, locality and size-extensivity.
The resulting model yields accurate CC-quality energies and forces, while simultaneously recovering dipoles, quadrupoles, polarizabilities, the electron density, and 2-electron observables such as the pair density.
We show that \molelambda{} further extends the speed advantage of \mole{} over full CCSD while substantially expanding the accessible properties, providing a route to wavefunction-level surrogate models for correlated quantum chemistry.
\end{abstract}

\section{Introduction}
\label{sec:introduction}

The ability to predict molecular properties accurately and efficiently is central to computational chemistry. Density functional theory (DFT) provides an attractive compromise between cost and accuracy and therefore underlies much of modern atomistic simulation \citep{kohn1965self,mardirossian2017thirty}. For applications that require substantially higher fidelity, however, correlated wavefunction methods remain essential \citep{szabo2012modern,helgaker2013molecular}. In particular, coupled-cluster theory is the standard route to chemically accurate ground-state properties, but its steep computational scaling of $O(N^6)$ limits routine use beyond small systems \citep{purvis1982full,raghavachari1989fifth,Bartlett2007RMP}.

\paragraph{Related work}
Recent machine learning work has narrowed this accuracy--efficiency gap from several directions. Machine-learned interatomic potentials can predict energies and forces at very low cost \citep{smith2020ani,mehdizadeh_surface_2025,martyka_omni-p2x_2025,burger2025hip,chen_one_2025,banchode_machine-learned_2025}. 
A separate line of work learns densities, density matrices, Hamiltonians, or Fock-like intermediates to improve one-particle observables or accelerate self-consistent electronic-structure calculations \citep{brockherde2017bypassing,shao2023machine,hazra2024predicting,febrer2025graph2mat,elsborg2025electra,jorgensen2022equivariant,yu2023efficient,luo2025efficient,kim2025high,li2025enhancing}.
More recent work has explored machine learning targets beyond standard DFT observables, including coupled-cluster properties, Green's functions, and wavefunction-inspired representations \citep{townsend2019data,townsend2020transferable,venturella_machine_2024,venturella_unified_2025,shang_solving_2025,gao_generalizing_2023,scherbela_towards_2024}. 

These approaches have been highly effective for their intended tasks, but they do not directly parameterize the coupled-cluster response state.
Learning energies and forces is attractive for molecular dynamics and screening, but does not recover the underlying correlated electronic state.
Learning densities or density matrices provides access to one-particle observables, while learning Hamiltonians can accelerate self-consistent field procedures.
Yet many chemically relevant quantities, including response densities, multipole moments, polarizabilities, and energy gradients, ultimately depend on the correlated wavefunction.

A particularly promising abstraction level is therefore the set of coupled-cluster amplitudes.
Molecular Orbital Learning (MoLe) \citep{thiede2026mole}, showed that learning the right-hand $T$-amplitudes directly from Hartree--Fock molecular orbitals yields accurate CC energies, accurate one-particle densities, and useful warm starts for CC solvers, especially in the low-data regime where expensive CC labels are scarce.
However, the right-hand amplitudes alone do not fully characterize the relaxed response properties of the CC state \citep{eriksen_lagrangian_2014,helgaker2013molecular}.
In Lagrangian formulations of CC theory, these properties additionally require the left-hand de-excitation amplitudes $\Lambda$.
This observation motivates the central idea of the present work. Once the $T$ and $\Lambda$ amplitudes are available, one can recover not only energies but also analytic gradients, multipoles, polarizabilities, coupled-cluster reduced density matrices, natural occupations, and two-particle observables from a single learned electronic object.

\paragraph{Present work}
In this work, we introduce \molelambda, an extension of \mole{} that predicts both $T$ and $\Lambda$ amplitudes from localized Hartree--Fock molecular orbitals.
Architecturally, \molelambda{} preserves the symmetry-aware orbital representation of \mole{} and augments it with $\Lambda_1$ and $\Lambda_2$ readouts that mirror the $T_1$ and $T_2$ heads.
Empirically, we show that \molelambda{} substantially broadens the practical utility of amplitude learning.
The model yields accurate correlated energies and forces competitive with strong MLIP baselines, while simultaneously recovering higher-order molecular properties that standard energy models cannot access directly.
In particular, we demonstrate accurate dipoles, quadrupoles, polarizabilities, $\Lambda$-state 1-RDMs and electron densities, as well as the 2-RDM-derived pair density.
We further show that \molelambda{} obtains the $T$ and $\Lambda$ amplitudes over two orders of magnitude faster than full CCSD + $\Lambda$-equations, thereby moving beyond property-specific surrogates toward a unified machine-learned representation of correlated wavefunction response.

\section{Background and Preliminaries}
\label{sec:theory}

\paragraph{Hartree--Fock\label{sec:lomo}}

We begin from restricted Hartree--Fock molecular orbitals (MOs) $\psi_p(\mathbf r \mid \{\mathbf R_A\})$. 
Each spatial orbital is expanded in an atomic-orbital basis as
\begin{equation}
\psi_p(\mathbf r)
= \sum_A
  \sum_{k \in \mathcal K_A}
  \sum_{\ell \in \mathcal L_{A,k}}
  \sum_{m=-\ell}^{\ell}
  C_{pA,k\ell m}\,
  \phi_{k\ell}^{m}(\mathbf r-\mathbf R_A),
\label{eq:mo-expansion}
\end{equation}
where $A$ labels atoms with positions $\mathbf R_A$, $k,\ell,m$ denote the principal, azimuthal, and magnetic quantum numbers, and $C_{pA,k\ell m}$ are elements of the MO coefficient matrix $\mathbf C$. Throughout the paper, $p,q$ index all orbitals, $i,j$ index occupied orbitals, and $a,b$ index virtual orbitals.
The MO coefficients are obtained by solving the Roothaan--Hall equations
\begin{equation}
\mathbf F(\mathbf C)\mathbf C = \mathbf S \mathbf C \bm{\varepsilon}
\label{eq:hf-scf}
\end{equation}
with $\mathbf F(\mathbf C)$ the Fock matrix, $\mathbf S$ the atomic-orbital overlap matrix, $\mathbf C$ the MO coefficient matrix, and $\bm{\varepsilon}$ the diagonal matrix of orbital energies. Since $\mathbf F$ depends on $\mathbf C$, this equation is solved iteratively.

\paragraph{Localized molecular orbitals} Canonical molecular orbitals are generally delocalized over the full molecule. We therefore work in a separately localized occupied and virtual gauge,
\begin{equation}
\tilde{\psi}_i = \sum_{j \in \mathrm{occ}} (U_{\mathrm{occ}})_{ji}\psi_j,
\qquad
\tilde{\psi}_a = \sum_{b \in \mathrm{virt}} (U_{\mathrm{virt}})_{ba}\psi_b,
\label{eq:localization}
\end{equation}
where $U_{\mathrm{occ}}$ and $U_{\mathrm{virt}}$ are unitary transformations within the occupied and virtual subspaces, respectively. This transformation leaves all observables invariant, while exposing the fragment structure of the amplitudes and thereby providing an important inductive bias for transfer across molecular size and chemistry. Unless noted otherwise, all subsequent orbital-indexed quantities are expressed in this localized MO basis.

\begin{figure*}[htbp]
    \centering
    \includegraphics[width=0.99\linewidth]{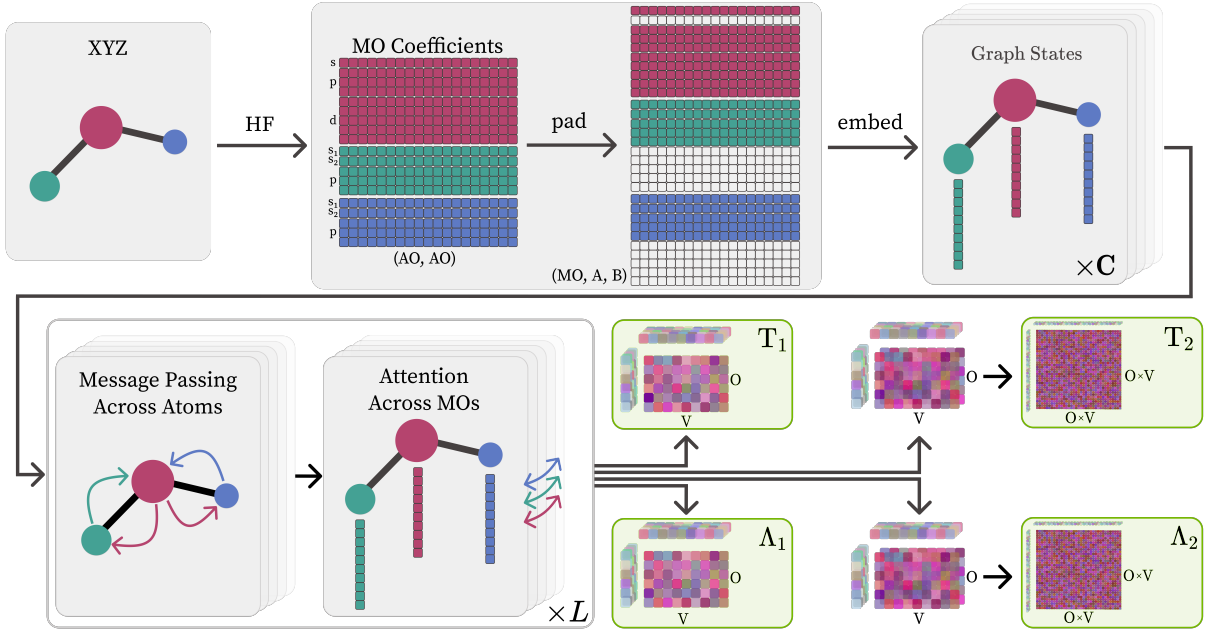}
    \caption{
    Given a molecule, a Hartree-Fock calculation provides the molecular orbitals (MOs) represented by their coefficient matrix $\mathbf{C}$. 
    The coefficient vector is padded for each atom to ensure that all atoms have the same number of basis coefficients, and embedded in an equivariant neural network. The model then alternates (a) message passing to mix information within the MOs and (b) attention layers to mix information between MOs. Finally, the embeddings are read out with specialized readout heads for the $(T_1,T_2,\Lambda_1,\Lambda_2)$ amplitudes.
    }
    \label{fig:model_overview}
\end{figure*}

\paragraph{Coupled-cluster right and left states}
In coupled-cluster theory, the correlated right-hand wavefunction is written as
\begin{equation}
\ket{\Psi_\text{CC}} = e^{\hat T}\ket{\Phi_{\text{HF}}},
\qquad
\hat T = \hat T_1 + \hat T_2,
\label{eq:right-state}
\end{equation}
where truncation to singles and doubles defines CCSD. Using standard spin-orbital notation for the second-quantized operators, with the restricted closed-shell implementation represented by the corresponding spin-adapted spatial-orbital tensors, the cluster operators are
\begin{align}
\hat T_1 = \sum_{ia} t_i^a\, \hat a_a^\dagger \hat a_i, \quad
\hat T_2 = \frac{1}{4}\sum_{ijab} t_{ij}^{ab}\, \hat a_a^\dagger \hat a_b^\dagger \hat a_j \hat a_i.
\label{eq:t-operators}
\end{align}
Here, $t_i^a$ and $t_{ij}^{ab}$ are the singles and doubles amplitudes. They are obtained from a nonlinear system of equations, and solving the CCSD equations formally scales as $\mathcal{O}(N^6)$ \citep{purvis1982full,szabo2012modern}. 
The corresponding correlation energy can be computed from only $T_1$ and $T_2$. For relaxed observables and response properties, the $T$ amplitudes alone are insufficient. 
Unlike variational methods, CC theory is not stationary with respect to the right amplitudes alone, so the total derivative $dE_{\mathrm{CC}}/d\xi$ contains extra terms $\partial t_\mu / \partial \xi$ for each perturbation $\xi$ and excitation $\mu$ that would need to be computed.
The coupled-cluster Lagrangian avoids this by introducing $\Lambda$ as an adjoint variable
\citep{koch1990Response}
\begin{equation}
\mathcal L(T,\Lambda)
=
\bra{\Phi_{\text{HF}}}
(1+\hat\Lambda)
e^{-\hat T}\hat H e^{\hat T}
\ket{\Phi_{\text{HF}}},
\label{eq:cc-lagrangian}
\end{equation}
with the adjoint left state $\bra{\widetilde\Psi_\text{CC}}$, and the de-excitation operators $\hat\Lambda_1, \hat\Lambda_2$
\begin{align}
\bra{\widetilde\Psi_\text{CC}}
=
\bra{\Phi_{\text{HF}}}(1+\hat\Lambda)e^{-\hat T},
\qquad
\hat\Lambda = \hat\Lambda_1 + \hat\Lambda_2
\\
\hat\Lambda_1 = \sum_{ia} \lambda_a^i\, \hat a_i^\dagger \hat a_a, \quad
\hat\Lambda_2 = \frac{1}{4}\sum_{ijab} \lambda_{ab}^{ij}\, \hat a_i^\dagger \hat a_j^\dagger \hat a_b \hat a_a
\label{eq:lambda-operators}
\end{align}
The introduction of $\Lambda$ simplifies the derivative of a non-variational CC energy into the partial derivative of a stationary Lagrangian \citep{Bartlett2007RMP}
\begin{equation}
\frac{dE}{d\xi}
=
\frac{\partial \mathcal L(T,\Lambda;\xi)}{\partial \xi},
\label{eq:lagrangian-derivative}
\end{equation}
where $T=\{t_i^a,t_{ij}^{ab}\}$ and $\Lambda=\{\lambda_a^i,\lambda_{ab}^{ij}\}$ denote the CCSD amplitude tensors, while $\hat T$ and $\hat\Lambda$ denote the corresponding second-quantized operators.
In conventional CCSD, the $T$ amplitudes have to be solved first, after which the $\Lambda$ amplitudes can be obtained from the linear adjoint equations
\begin{equation}
    \frac{\partial \mathcal L}{\partial t_\mu}
    =
    \frac{\partial E_{\mathrm{CC}}}{\partial t_\mu}
    +
    \sum_\nu
    \lambda_\nu
    \frac{\partial R_\nu}{\partial t_\mu}
    =
    0 .
\end{equation}
Unfortunately, these CCSD $\Lambda$-equations are dominated by the same type of four-virtual/two-occupied contractions that dominate CCSD, with a scaling of $\mathcal{O}(N^6)$.
If the amplitudes $(T,\Lambda)$ can be made accessible cheaply, however, they form the \ccsd{} response-state required for coupled-cluster density matrices, analytic gradients, and response properties, making them an attractive learning target.

\section{M\=oLe-$\mathbf{\Lambda}$ \label{sec:method}}

\begin{table*}[htbp]
\centering
\footnotesize
\setlength{\tabcolsep}{4pt}
\renewcommand{\arraystretch}{0.95}
%
\begin{tabular*}{\textwidth}{@{\extracolsep{\fill}}lcccccccccccc}
\toprule
& \multicolumn{2}{c}{Validation}& \multicolumn{4}{c}{Size extrapolation}& \multicolumn{6}{c}{Out-of-equilibrium} \\
\cmidrule(lr){2-3}
\cmidrule(lr){4-7}
\cmidrule(lr){8-13}
  & \multicolumn{2}{c}{QM7} & \multicolumn{2}{c}{Amino acids} & \multicolumn{2}{c}{PubChem} & \multicolumn{2}{c}{Diels-Alder} & \multicolumn{2}{c}{Dihedral scan} & \multicolumn{2}{c}{Chair-to-boat} \\
\cmidrule(lr){2-3}\cmidrule(lr){4-5}\cmidrule(lr){6-7}\cmidrule(lr){8-9}\cmidrule(lr){10-11}\cmidrule(lr){12-13}
  & E & F & E & F & E & F & E & F & E & F & E & F \\
\midrule
MP2 & 57.32 & 1.50 & 60.49 & 1.33 & 82.55 & \underline{1.32} & 69.33 & \underline{1.18} & 60.99 & 0.57 & 75.03 & 0.54 \\
\midrule
Mace & 0.79 & 1.20 & 9.03 & 9.99 & 19.45 & 9.44 & 11.25 & 7.99 & 2.92 & 1.03 & 3.19 & 1.87 \\
Mace+MP2 & 0.16 & 0.23 & \underline{0.51} & 1.90 & \underline{2.07} & 2.49 & \underline{1.61} & 1.43 & \underline{0.34} & 0.90 & \underline{0.41} & 0.83 \\
eSEN & 0.74 & 0.89 & 11.26 & 6.70 & 31.53 & 7.54 & 16.39 & 12.08 & 2.14 & 7.07 & 2.89 & 5.43 \\
eSEN+MP2 & \underline{0.15} & \underline{0.17} & 3.20 & \underline{0.69} & 8.12 & 1.81 & 1.81 & 1.94 & 0.63 & \underline{0.17} & 0.68 & \underline{0.19} \\
\midrule
M\=oLe-$\Lambda$ & \textbf{0.10} & \textbf{0.12} & \textbf{0.37} & \textbf{0.27} & \textbf{0.63} & \textbf{0.26} & \textbf{1.09} & \textbf{0.24} & \textbf{0.29} & \textbf{0.12} & \textbf{0.40} & \textbf{0.13} \\
\bottomrule
\end{tabular*}
\normalsize
\caption{
Mean absolute error of energy and force predictions in mHa and mHa/Bohr.
}
\label{tab:results}
\end{table*}

We now present \molelambda{}, which extends Molecular Orbital Learning (\mole{}) \citep{thiede2026mole} to predict both the right-hand coupled-cluster amplitudes
$
T = (T_1, T_2)
$
and the left-hand adjoint amplitudes
$
\Lambda = (\Lambda_1, \Lambda_2)
$
directly from localized Hartree--Fock molecular orbitals.
Our architecture design follows the same philosophy as \mole{}, now applied to the full response state. The model should satisfy four constraints:
\begin{enumerate}
    \item Molecular-orbital coefficients are rotation equivariant
    \item Predicted amplitudes are rotationally invariant
    \item Predicted amplitudes are sign equivariant with respect to orbital phase flips
    \item Amplitudes coupling localized orbitals on non-interacting fragments should vanish
\end{enumerate}

With these constraints in mind, \molelambda{} first runs a fast HF calculation to obtain MOs, performs orbital localization, passes the localized MOs into a shared equivariant encoder and backbone, and finally predicts $(T_1,T_2,\Lambda_1,\Lambda_2)$ via separate readout heads, as depicted in Fig.~\ref{fig:model_overview}.
The output amplitudes are subsequently passed to conventional coupled-cluster post-processing routines to obtain energies, forces, and response observables.

\paragraph{Shared equivariant encoder}
We reuse the orbital-centric representation of \mole{}. Each localized orbital is embedded into an equivariant latent representation, one graph per MO, and processed by the same backbone of molecular-orbital attention, Odd-MACE updates, and equivariant normalization \citep{qin_devil_2022,batatia_mace_2022,thiede2026mole}. The encoder is shared across occupied and virtual orbitals and across all four readout heads. 

\paragraph{Amplitude readout}

The original \mole{} readout produces singles amplitudes from pairwise invariant features $\mathbf y_{ia}$ and doubles amplitudes from four-orbital invariant features $\mathbf y_{ijab}$. We retain that structure for the $T$ amplitudes, and mirror it for the left state,
\begin{align}
t_i^a = \mathrm{OddReadout}_{T_1}(\mathbf y_{ia}), 
\\ t_{ij}^{ab} = \mathrm{OddReadout}_{T_2}(\mathbf y_{ijab})
\\
\label{eq:readout}
\lambda_a^i = \mathrm{OddReadout}_{\Lambda_1}(\mathbf y_{ia}), 
\\ 
\lambda_{ab}^{ij} = \mathrm{OddReadout}_{\Lambda_2}(\mathbf y_{ijab})
\end{align}
The odd readout ensures sign equivariance, while the invariant features guarantee rotational invariance of the predicted amplitudes. 

\begin{figure*}[htbp] 
    \centering
    \includegraphics[width=0.99\linewidth]{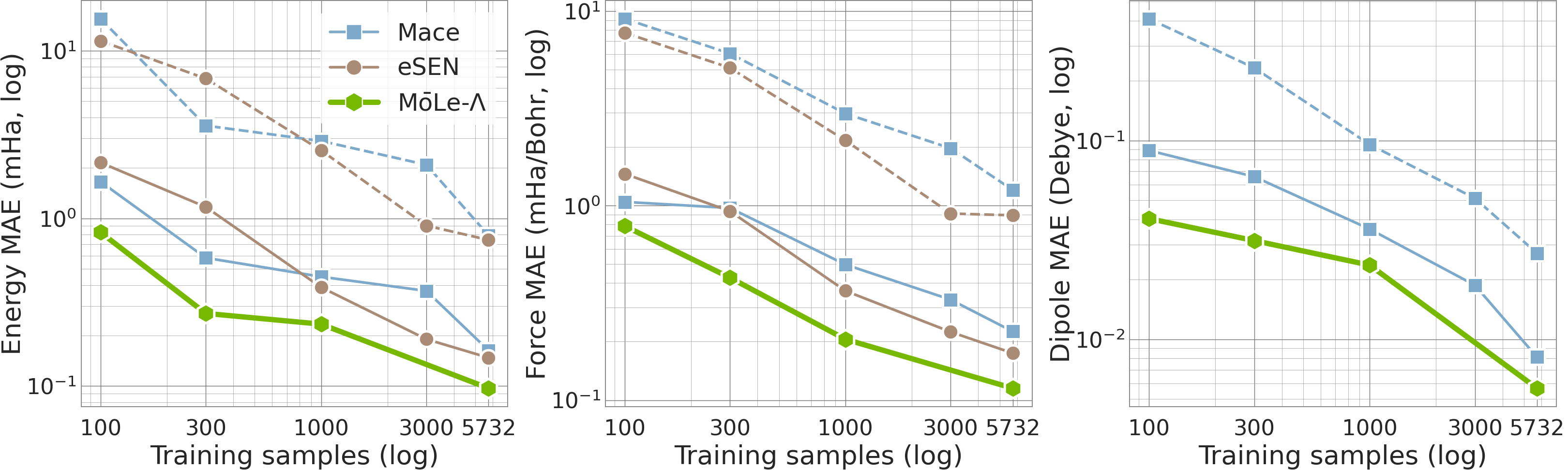}
    \caption{Scaling of the energy, force, and dipole MAE of \molelambda{} and MLIPs with increasing data. Dashed lines predict CCSD labels, solid lines predict the delta between CCSD and MP2.}
    \label{fig:datascaling}
\end{figure*}

\paragraph{Direct and residual targets}
In the present restricted closed-shell systems with real amplitudes, the lowest-order perturbative structure provides a natural prior.
For the right-hand doubles amplitudes $T_2$, the natural low-order baseline is the canonical-orbital MP2 amplitude
\begin{align}
    t_{ij,\mathrm{MP2}}^{ab}
    =
    \frac{\langle ij || ab \rangle}
    {\varepsilon_i + \varepsilon_j - \varepsilon_a - \varepsilon_b},
\end{align}
with $T_1^{\mathrm{MP2}} = 0$. For the left-hand amplitudes, restricted closed-shell systems with real amplitudes admit an analogous perturbative simplification:
\begin{equation}
\Lambda_1^{\mathrm{MP2}} = 0,
\qquad
\Lambda_2^{\mathrm{MP2}} = {T_2^{\mathrm{MP2}}}^* = T_2^{\mathrm{MP2}}
\label{eq:mp2-relations}
\end{equation}
For residual training, these MP2 amplitudes are computed in the canonical HF basis and then transformed with the same occupied and virtual localization matrices as the CCSD amplitudes. After this transformation, we use the same symbols for the localized-gauge tensors.
This gives two natural training strategies for the $\Lambda$ heads.
In direct mode, we predict $t_{ij}^{ab}$ and $\lambda_{ab}^{ij}$ outright.
In residual mode, we instead learn the corrections
\begin{align}
    \Delta t_{ij}^{ab}
    &=
    t_{ij,\mathrm{CCSD}}^{ab} - t_{ij,\mathrm{MP2}}^{ab}, 
    \\
    \Delta \lambda_{ab}^{ij}
    =
    \lambda_{ab,\mathrm{CCSD}}^{ij} &- \lambda_{ab,\mathrm{MP2}}^{ij},
    \quad
    \lambda_{ab,\mathrm{MP2}}^{ij} \equiv t_{ij,\mathrm{MP2}}^{ab}.
\end{align}
Residualization is especially attractive in the low-data regime, because it offloads the leading-order correlation structure to perturbation theory and asks the network to learn only the physically smaller, but chemically crucial, correction.

\paragraph{Training objective}
The training objective is an amplitude reconstruction loss implemented as a per-molecule sum of elementwise squared amplitude errors, averaged over the batch,
\begin{equation}
\mathcal J_{\mathrm{amp}}
=
\frac{1}{B}
\sum_{b=1}^{B}
\sum_{\substack{X \in \\ \{T_1,T_2,\Lambda_1,\Lambda_2\}}}
w_X \sum_{n=1}^{N_X^{(b)}}
\left(\hat X_{b,n} - X^{\mathrm{ref}}_{b,n}\right)^2 ,
\label{eq:amp-loss}
\end{equation}
where $B$ is the batch size and $N_X^{(b)}$ denotes the number of supervised elements in tensor $X$ for molecule $b$. When residual targets are used, the same loss is applied to the residual tensors instead of the raw amplitudes.
The loss weights $w_X$ can be tuned for optimal downstream performance, and we set all active amplitude loss weights to one for simplicity.

An important feature of this training objective is that we do not supervise individual downstream properties with separate prediction heads. All reported energies, forces, multipoles, densities, and two-electron observables are obtained from the predicted amplitudes. Instead of learning each property independently, the model learns the underlying coupled-cluster representation.

\paragraph{Coupled-cluster property reconstruction}

The CCSD correlation energy depends only on the right-hand amplitudes and is obtained from the predicted $T_1$ and $T_2$ amplitude tensors \citep{Bartlett2007RMP}
\begin{align}
    E_{\mathrm{corr}}
    &=
    \sum_{ijab}
    \left(
        \frac{1}{4} t_{ij}^{ab}
        +
        \frac{1}{2} t_i^a t_j^b
    \right)
    \langle ij || ab \rangle
    , \\
    E_{CC} &= E_{HF} + E_{corr}
\end{align}

To access relaxed properties we require the coupled-cluster Lagrangian in eq.~\ref{eq:cc-lagrangian},
which allows us to compute analytical derivatives with respect to an external perturbation $\xi$ via \ref{eq:lagrangian-derivative}.
Forces are computed from the derivative with respect to nuclear coordinates
\begin{align}
    \mathbf F_A
    =
    -
    \frac{\partial \mathcal{L}(T,\Lambda)}
    {\partial \mathbf R_A}
\end{align}
To account for the response of the molecular orbitals themselves under nuclear perturbations, one constructs CCSD density intermediates from $(T,\Lambda)$ and augments them with orbital-response contributions obtained by solving a coupled perturbed Hartree--Fock (CPHF) equation \citep{adamowicz1984analytical, handy1984evaluation}.

A large class of fixed-orbital observables can be computed directly as expectation values of the $\Lambda$-state one- and two-particle reduced density matrices (1-/2-RDMs) \citep{Stanton1993EOMCC,Bartlett2007RMP}.
From the predicted amplitudes, we reconstruct the fixed-orbital CCSD 1-RDM and 2-RDM via
\begin{align}
    \gamma_{pq}
    &=
    \bra{\Phi_0}(1+\hat \Lambda)e^{-\hat T}
    a_p^\dagger a_q
    e^{\hat T}\ket{\Phi_0}, \\
    \Gamma_{pq,rs}
    &=
    \bra{\Phi_0}(1+\hat \Lambda)e^{-\hat T}
    a_p^\dagger a_q^\dagger a_s a_r
    e^{\hat T}\ket{\Phi_0}
\end{align}
This convention defines $\gamma_{pq}$ and $\Gamma_{pq,rs}$ with the same operator ordering that appears in the contractions below.
For any one-body operator $\hat O^{(1)}$, its expectation value can be obtained from the 1-RDM
\begin{equation}
\hat O^{(1)}=\sum_{pq}o_{pq}a_p^\dagger a_q, \quad
    \langle \hat O^{(1)} \rangle
    =
    \sum_{pq} o_{pq}\gamma_{pq}
\end{equation}
This includes the electron density, dipole moment, quadrupole moment, and, in principle, higher multipoles.
For any two-body operator $\hat O^{(2)}$ the expectation value uses the 2-RDM
\begin{align}
\hat O^{(2)}&=\frac14\sum_{pqrs}v_{pq,rs}a_p^\dagger a_q^\dagger a_s a_r, 
    \\
    \langle \hat O^{(2)} \rangle
    &=
    \frac14
    \sum_{pqrs}
    v_{pq,rs}\Gamma_{pq,rs}
\end{align}
In real space, the 2-RDM determines quantities such as the pair density.
Field-response quantities such as polarizabilities are obtained by evaluating the reconstructed reduced density matrices under weak external perturbations and forming the corresponding derivatives (see Sec.~\ref{si:polarizability}).

\section{Experiments\label{sec:experiments}}

\begin{figure*}[htbp]
    \centering
    \includegraphics[width=0.99\linewidth]{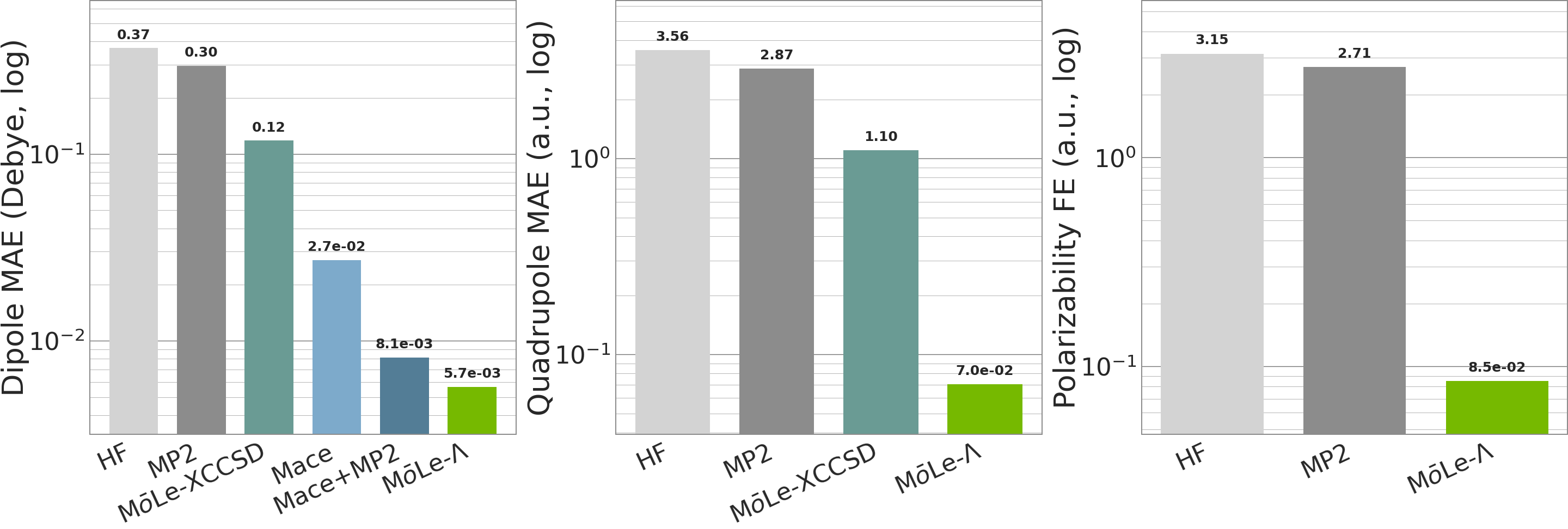}
    \caption{
    Mean absolute error for dipoles, quadrupoles, and polarizability on the QM7 test set. Plotted on a log scale.
    }
    \label{fig:barchart}
\end{figure*}

We evaluate whether a learned coupled-cluster response state can replace the explicit solution of the CCSD and $\Lambda$ equations for downstream property reconstruction.
The central object is the set of four amplitude tensors
$
(T_1,T_2,\Lambda_1,\Lambda_2)
$.
Once these amplitudes are predicted, all reported energies, gradients, multipoles, density matrices, densities, and pair-density observables are obtained by standard coupled-cluster post-processing rather than training property-specific prediction heads.

To train \molelambda{} we recalculate the QM7 dataset using coupled-cluster. QM7 \citep{rupp,blum} contains 7165 small organic molecules composed of C, N, O, S, and H, which we divide into a 80/20 train--test split.
For each geometry in the dataset, we compute restricted Hartree--Fock orbitals and CCSD response amplitudes
$
(T_1,T_2,\Lambda_1,\Lambda_2)
$
at the CCSD/def2-SVP level of theory \citep{weigend2005balanced}.
All amplitudes are transformed to the separately localized occupied and virtual orbital gauge used by the model.
Training details and hyperparameters are given in Appendix~\ref{si:hyperparameters_neurips_lambda}.

\paragraph{Baselines}
We compare against Hartree--Fock, MP2, and, where appropriate, \mole{} with XCCSD-style reconstructions from predicted $T$ amplitudes only.
HF provides the uncorrelated mean-field reference.
MP2 provides a perturbative correlated baseline and, in residual mode, the low-order doubles amplitudes used by \molelambda{}.
However, MP2 does not provide the full relaxed CCSD response state.
Thus, improvements over MP2 for $\Lambda$-dependent observables measure whether the model has learned corrections beyond leading-order perturbation theory.
For energies and forces, we additionally compare against machine-learned interatomic potentials trained on the same data, as described in Appendix~\ref{si:mlip}.

\paragraph{Coupled-cluster energies and gradients}
\label{sec:results-energies}

We first assess whether the predicted amplitudes reproduce the CCSD right and left states.
On the QM7 test set, \molelambda{} achieves elementwise mean absolute errors of
$
2.6\text{--}2.7\times10^{-5}
$
for $T_1$ and $\Lambda_1$, and
$
5.3\times10^{-7}
$
for both $T_2$ and $\Lambda_2$.
These amplitude errors translate into small downstream errors in the CCSD correlation energy and analytic forces, with MAEs of
$
0.10
$
mHa and
$
0.12
$
mHa/Bohr, respectively.
As shown in Table~\ref{tab:results}, \molelambda{} improves over both direct CCSD MLIP baselines and $\Delta$-MP2 MLIP baselines.

We next test transfer to larger molecules.
The first size-extrapolation set contains 18 structurally diverse amino acids with up to 15 heavy atoms, roughly twice the size of the largest QM7 training molecules.
The second contains 100 PubChem molecules with 14 heavy atoms, spanning diverse scaffolds, functional groups, ring systems, and heterocycles.
On both sets, \molelambda{} retains low CCSD energy and force errors, whereas the geometry-only MLIP baselines degrade more strongly.
This supports the hypothesis that localized orbital amplitudes provide a size-transferable representation of correlation.

Finally, we probe out-of-equilibrium generalization on three chemically distinct scans: the Diels--Alder reaction between ethylene and 1,3-butadiene, a butane central-bond dihedral scan, and the chair-to-boat conformational change of cyclohexane.
Across all three scans, \molelambda{} gives the lowest or tied-lowest energy errors and the lowest force errors in Table~\ref{tab:results}, indicating that the orbital response-state surrogate extrapolates more robustly than property-specific MLIPs.

\paragraph{Data efficiency and scaling}

Because CCSD labels are expensive, the surrogate must be accurate in the low-data regime.
We therefore train \molelambda{} and the MLIP baselines on increasingly large subsets of QM7 and evaluate the resulting energy, force, and dipole errors.
Fig.~\ref{fig:datascaling} shows that \molelambda{} consistently improves over MLIPs at all tested data regimes, suggesting that supervising the structured amplitude tensors
$
(T_1,T_2,\Lambda_1,\Lambda_2)
$
is more data-efficient than directly fitting molecular properties.

\begin{figure}[!ht]
    \centering
    \includegraphics[width=0.98\linewidth]{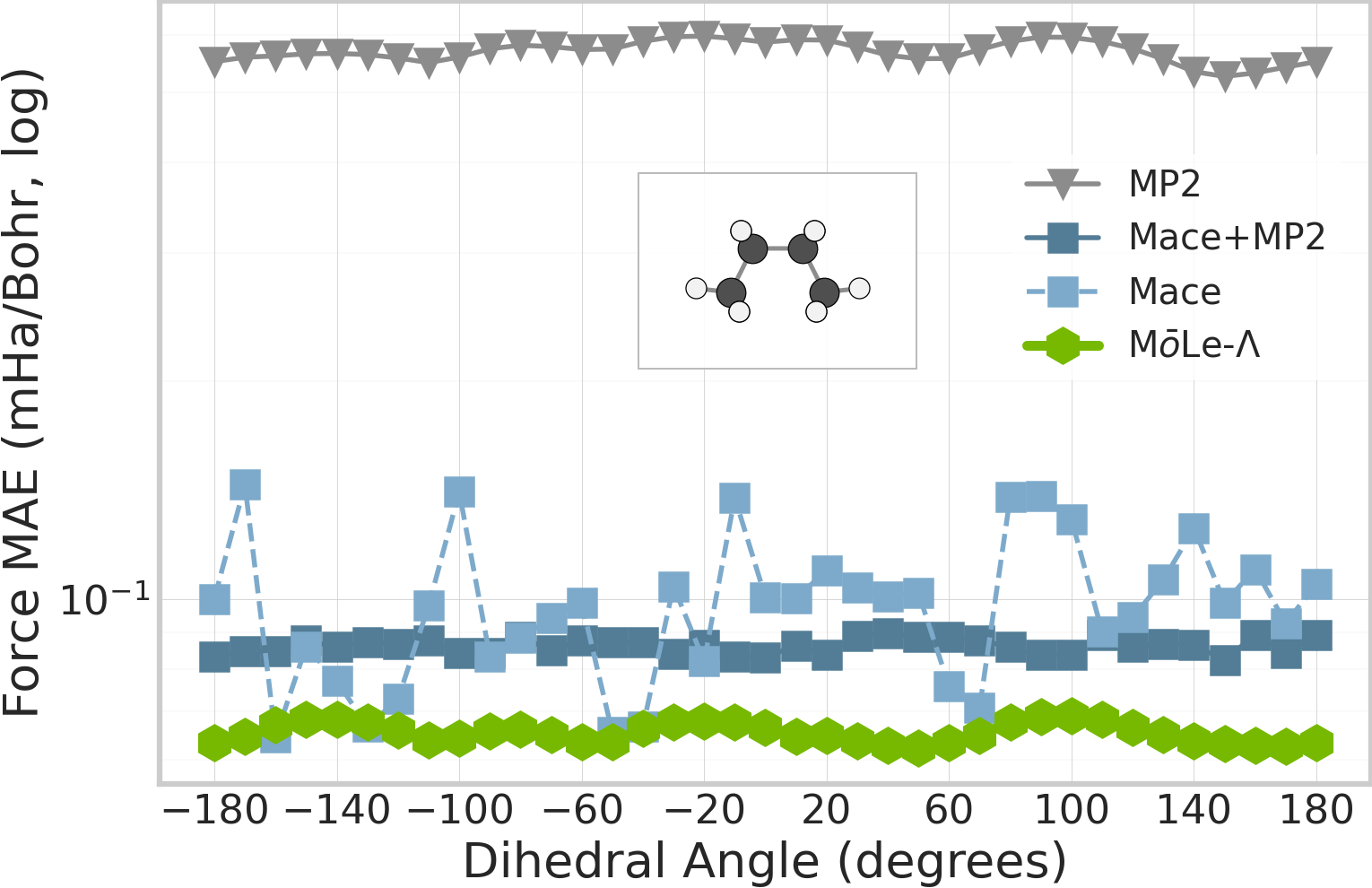}
    \includegraphics[width=0.98\linewidth]{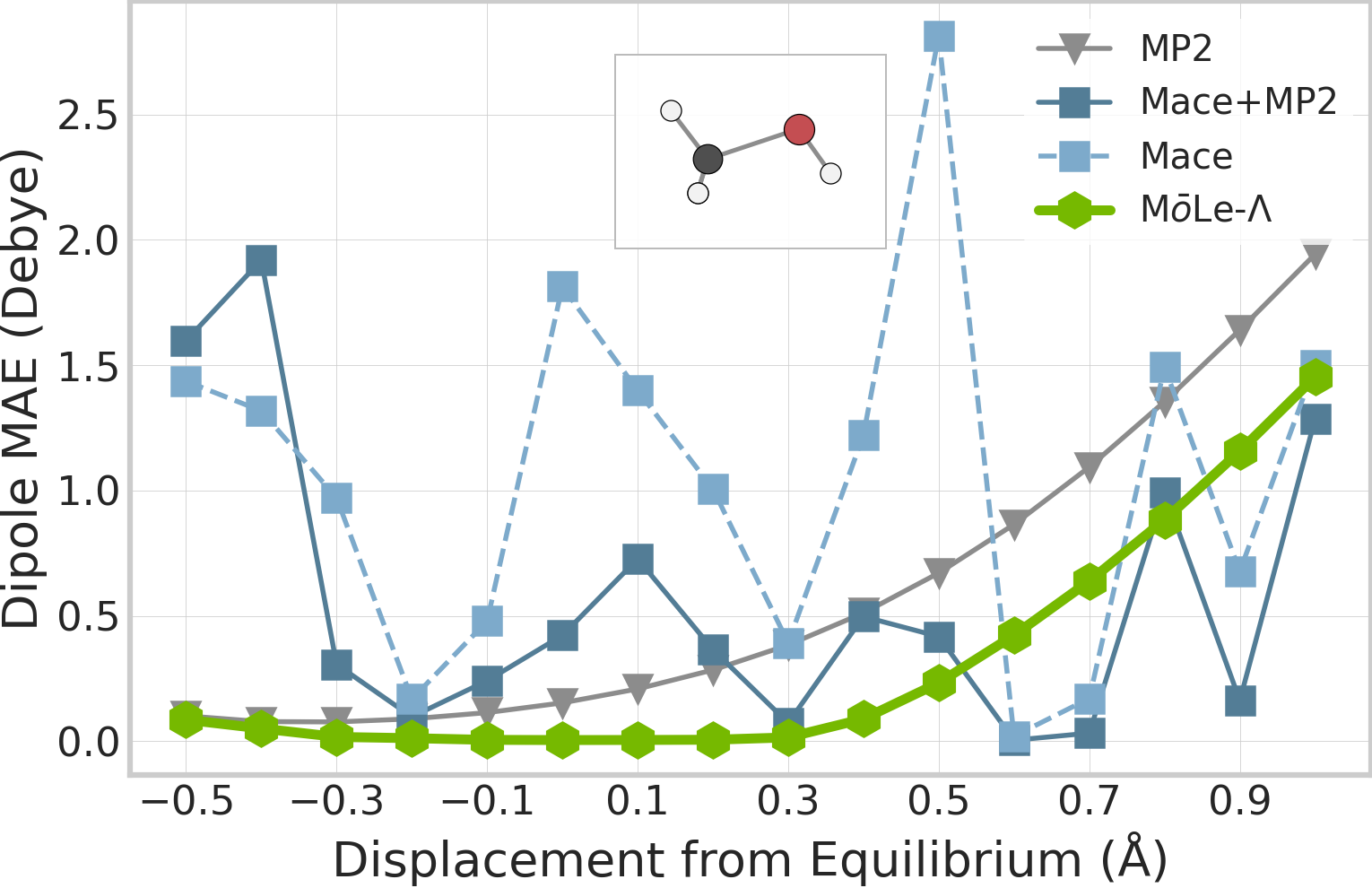}
    \caption{Out-of-equilibrium tests of the force MAE along a dihedral scan of the centre bond in a butane molecule (top) and the dipole MAE of methanol stretched along the C--O bond (bottom).}
    \label{fig:ooe}
\end{figure}

\paragraph{Multipoles and static response}
\label{sec:results-response}

We next evaluate observables that require the coupled-cluster response state, namely dipoles, quadrupoles, and polarizability.
As shown in Fig.~\ref{fig:barchart}, \molelambda{} substantially improves over HF, MP2, and right-state-only \mole{}-XCCSD reconstructions.
This indicates that the predicted $\Lambda$ amplitudes provide the adjoint information needed for accurate one-particle and response observables.

Figure~\ref{fig:ooe} reports the force MAE along the central C--C dihedral scan of butane (top) and the dipole MAE of methanol stretched along the C--O bond (bottom).
The Mace model trained directly on CCSD data gives wildly inconsistent predictions. The $\Delta$-MP2 Mace performs better on the dihedral scan, but still worse than our method, and shows erratic dipole errors on the bond stretch. 
\molelambda{} consistently improves over MP2 and both baselines. 
This suggests that learning the response state in an orbital representation provides a more stable extrapolation mechanism than fitting properties alone.

\begin{figure}[!htbp]
    \centering
    \includegraphics[width=0.99\linewidth]{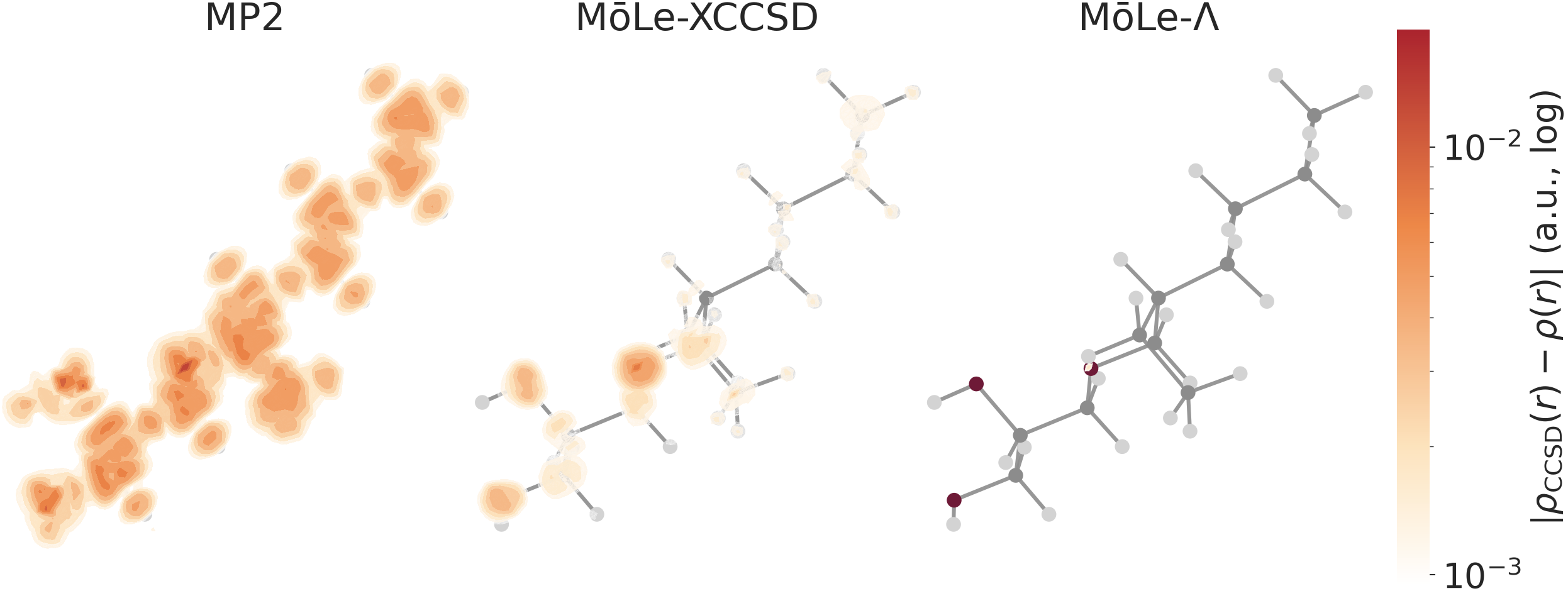}
    \caption{
    Difference in the electron density $\rho(\mathbf{r})$ of MP2, \mole-XCCSD, and \molelambda{} compared to CCSD on ethylhexylglycerin, PubChem CID 9859093, $C_{11}H_{24}O_{3}$.
    Oxygen is coloured red, nitrogen blue, carbon dark grey, and hydrogen light grey.
    }
    \label{fig:density}
\end{figure}

\begin{figure}[!htbp]
    \centering
    \includegraphics[width=0.99\linewidth]{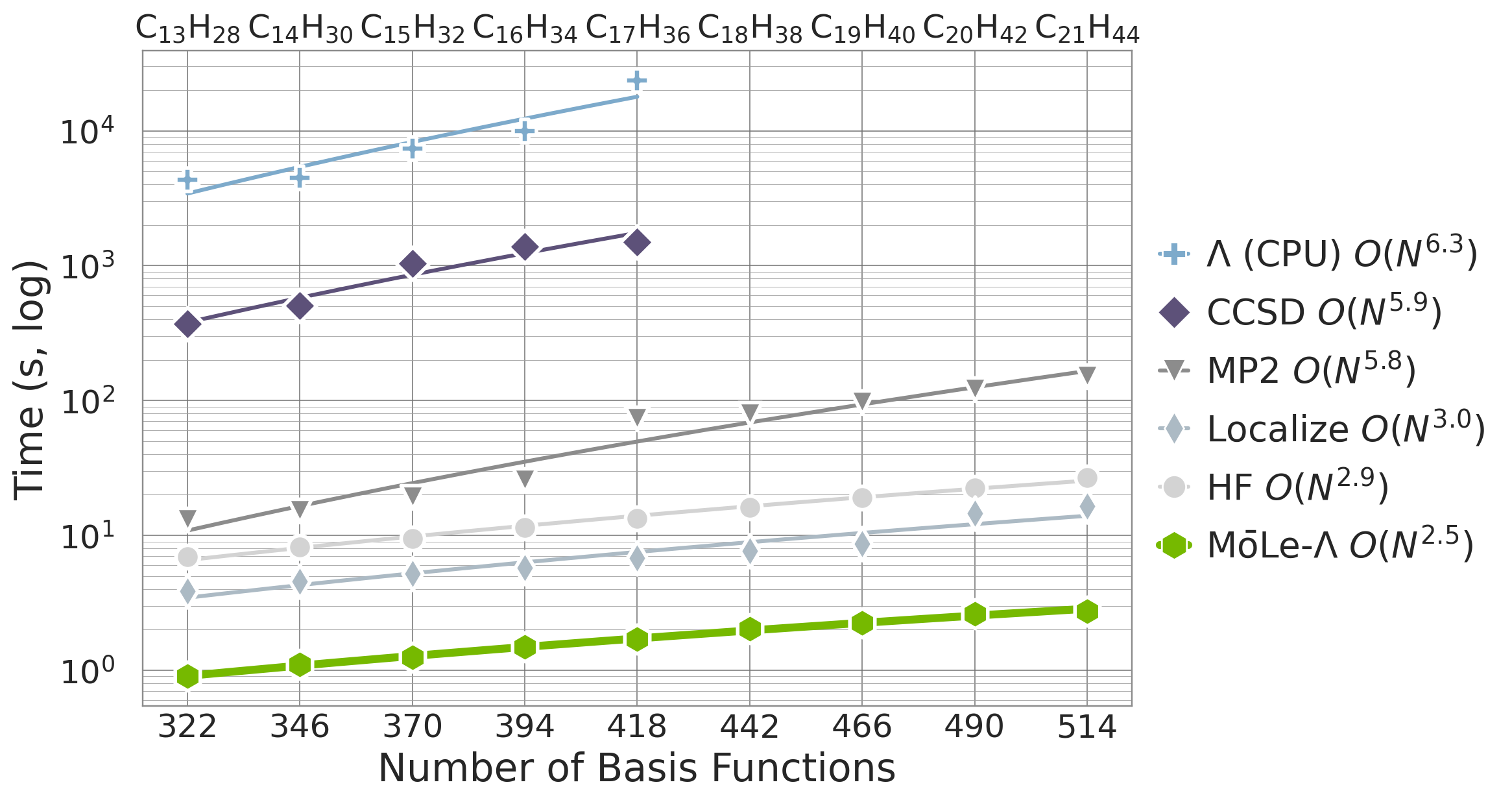}
    \caption{Empirical scaling timed on a H100 for alkane chains of increasing size. CCSD, MP2, and HF use GPU4PySCF \citep{li2025gpu4pyscf}. The $\Lambda$-equations and localization, are timed on CPU.}
    \label{fig:scaling}
\end{figure}

\paragraph{Electron density via the 1-RDM}
\label{sec:results-1rdm}

We then evaluate the real-space electron density reconstructed from the $\Lambda$-state 1-RDM.
Figure~\ref{fig:density} compares the density error
for MP2, right-state-only \mole{} with XCCSD-style reconstruction, and \molelambda{}.
MP2 produces a broad and diffuse error over the molecular volume, indicating an extended deviation from the CCSD density.
\mole{} significantly reduces both the magnitude and spatial extent of the residual relative to CCSD. \molelambda{} further suppresses off-atom and bond-adjacent density errors.
This improvement demonstrates the practical value of learning the adjoint amplitudes rather than the right state alone.

\paragraph{Pair density via the 2-RDM}
\label{sec:results-2rdm}

\begin{figure}
    \centering
    \includegraphics[width=0.9\linewidth]{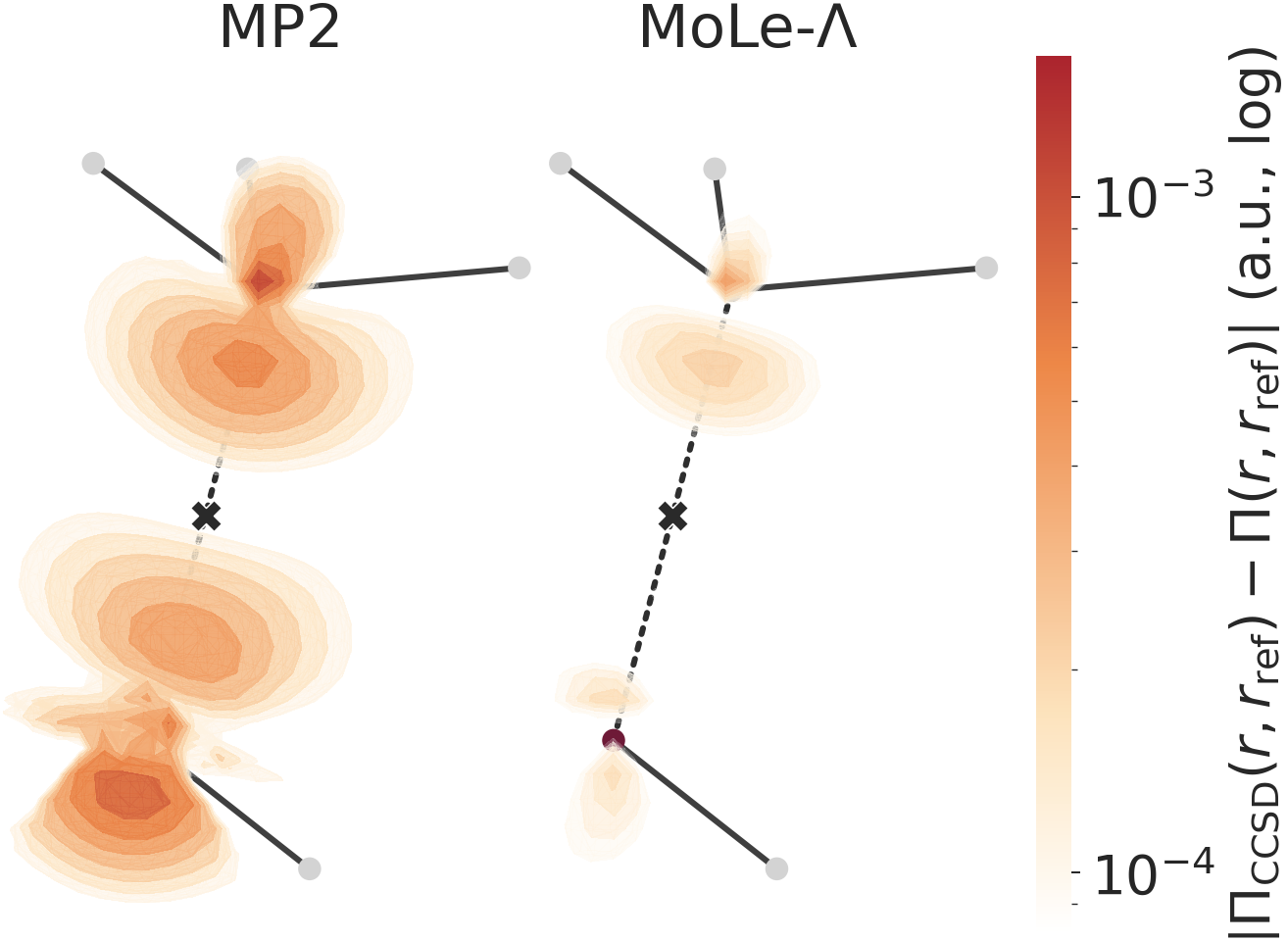}
    \caption{
    Difference in the pair-density $\Pi(\mathbf{r},\mathbf{r}_{\mathrm{ref}})$, obtained from the 2-RDM, of MP2 and \molelambda{} relative to CCSD on a methanol geometry stretched along the C--O bond.
    }
    \label{fig:pairdensity}
\end{figure}

Unlike energies, forces, or one-body properties, the pair density 
$
\Pi(\mathbf r,\mathbf r_{\mathrm{ref}})
$
depends directly on the $\Lambda$-state 2-RDM and therefore probes whether \molelambda{} captures correlated electron-pair structure.
Figure~\ref{fig:pairdensity} compares the pair-density error
for stretched methanol along the C--O bond, with the reference point $\mathbf r_{\mathrm{ref}}$ placed in the internuclear region.
Relative to MP2, \molelambda{} yields a smaller and more localized error.
The broad MP2 error lobes near the bond are strongly suppressed, indicating that the learned response state better captures the redistribution of electron-pair correlation in the stretched polar bond.

\paragraph{Computational scaling}
Finally, we compare wall-clock scaling against conventional CCSD+$\Lambda$ calculations. 
Figure~\ref{fig:scaling} reports timings for increasing alkane chains using GPU4PySCF \citep{li2025gpu4pyscf}; orbital localization and the $\Lambda$ equations are only available on CPU. 
Conventional CCSD calculations include HF, localization, MP2, CCSD, and the $\Lambda$ solve, whereas our method replaces the CCSD+$\Lambda$ steps by optional MP2 calculation and a \molelambda{} forward pass.
For \molelambda{}, the dominant outputs are the dense doubles tensors $T_2$ and $\Lambda_2$, each of size $\mathcal O(N_{\mathrm{occ}}^2N_{\mathrm{virt}}^2)$. 
The empirical scaling of \molelambda{} is lower than $\mathcal O(N^3)$ in the tested regime, likely because the dominant tensor operations are efficiently parallelized on GPU.
At $C_{17}H_{36}$, CCSD exhausts H100 memory, while \molelambda{} scales far beyond $C_{21}$. 
At the largest common system size, \molelambda{} obtains both left- and right-hand amplitudes more than two orders of magnitude faster than solving the CCSD and $\Lambda$ equations. 
The dense $T_2$ and $\Lambda_2$ outputs are not yet the practical bottleneck in the present experiments, but sparse, local, or compressed doubles representations provide a natural path to larger systems.
Further gains should follow from GPU-native localization and MP2.

\paragraph{Limitations and future work}
The present study is restricted to the def2-SVP basis and to the elements represented in QM7.
Larger basis sets, heavier elements, open-shell systems, and more diverse chemical environments are natural next steps.
The current implementation also treats HF, localization, and optional MP2 residual construction as external preprocessing steps; these are not yet optimized to the same degree as the neural network evaluation.
Finally, future work should explore sparse representations for $T_2$ and $\Lambda_2$ tensors to scale to substantially larger systems.

\section{Conclusions}
\label{sec:conclusion}
\molelambda{} extends orbital-based amplitude learning from the \ccsd{} right state to the full coupled-cluster ground-state response state. The core claim is that if a model predicts $(T_1,T_2,\Lambda_1,\Lambda_2)$ from localized Hartree--Fock molecular orbitals, then one learned object can support energies, forces, response properties, and reduced-density-matrix observables within a common framework. This reframes machine learning in correlated chemistry from approximating isolated observables to approximating the underlying electronic-structure state.

\section*{Acknowledgements}
A.A. gratefully acknowledged King Abdullah University of Science and Technology (KAUST) for the KAUST Ibn Rushd Postdoctoral Fellowship. 
A.B. and L.T. acknowledge the AIST support to the Matter lab for the project titled "SIP project - Quantum Computing". 
J.A.C.-G.-A. acknowledges funding of this project by the National Sciences and Engineering Research Council of Canada (NSERC) Alliance Grant \#ALLRP587593-23 (Quantamole) and also acknowledges support from the Council for Science, Technology and Innovation (CSTI), Cross-ministerial Strategic Innovation Promotion Program (SIP), “Promoting the application of advanced quantum technology platforms to social issues” (Funding agency: QST). 
A.A.-G. thanks Anders G. Fr{\o}seth for his generous support. A.A.-G. also acknowledges the generous support of Natural Resources Canada and the Canada 150 Research Chairs program. 
Resources used in preparing this research were provided, in part, by the Province of Ontario, the Government of Canada through CIFAR, and companies sponsoring the Vector Institute.
This research is part of the University of Toronto’s Acceleration Consortium, which receives funding from the CFREF-2022-00042 Canada First Research Excellence Fund. This research was enabled in part by support provided by SciNet and the Digital Research Alliance of Canada (\url{alliancecan.ca}).

\section*{Impact Statement}
This paper presents work whose goal is to advance machine learning methods for correlated quantum chemistry.
By making coupled-cluster-level response properties more accessible, \molelambda{} could reduce the computational cost of molecular simulation and support downstream research in materials, catalysis, energy, and molecular design.
As with other learned surrogates for scientific computing, possible negative impacts arise primarily from misuse or overinterpretation outside the training domain, where inaccurate energies, forces, or response properties could lead to misleading scientific conclusions.
The broader ethical and societal implications are otherwise those associated with advancing machine learning and computational chemistry; we do not identify additional impacts requiring specific emphasis.

\bibliography{references}
\bibliographystyle{icml2026}

\newpage
\appendix
\onecolumn

\section{Additional Background}

\subsection{$\Lambda$ amplitudes\label{si:lambda}}
In conventional CCSD, the right amplitudes are obtained first from the projected residual equations
$
R_\mu(T)=0
$,
where $\mu$ indexes single and double excitations.
For fixed converged $T$, the $\Lambda$ amplitudes are then obtained from the linear adjoint equations
\begin{equation}
    \frac{\partial \mathcal L}{\partial t_\mu}
    =
    \frac{\partial E_{\mathrm{CC}}}{\partial t_\mu}
    +
    \sum_\nu
    \lambda_\nu
    \frac{\partial R_\nu}{\partial t_\mu}
    =
    0 .
\end{equation}
$\Lambda$ acts as the Lagrange multiplier enforcing stationarity of the CC residuals, or equivalently the left eigenvector-like response state associated with the CC Jacobian.

The reason $\Lambda$ is useful can be seen from differentiating the CC energy with respect to an external perturbation $\xi$.
Since the converged amplitudes depend on the perturbation, $T=T(\xi)$, the total derivative contains an explicit term and an implicit amplitude-response term,
\begin{equation}
    \frac{dE_{\mathrm{CC}}}{d\xi}
    =
    \frac{\partial E_{\mathrm{CC}}}{\partial \xi}
    +
    \sum_\mu
    \frac{\partial E_{\mathrm{CC}}}{\partial t_\mu}
    \frac{\partial t_\mu}{\partial \xi},
\end{equation}
where $\mu$ indexes the single and double excitation amplitudes.
Unlike variational methods, CC theory is not stationary with respect to the right amplitudes alone, so the second term does not vanish in general.
Computing it directly would require solving for the response amplitudes
$
\partial t_\mu / \partial \xi
$
for each perturbation.

The Lagrangian avoids this by introducing $\Lambda$ as an adjoint variable.
Writing the projected CC residual equations as
$
R_\mu(T;\xi)=0
$,
the Lagrangian can be written schematically as
\begin{equation}
    \mathcal L(T,\Lambda;\xi)
    =
    E_{\mathrm{CC}}(T;\xi)
    +
    \sum_\mu \lambda_\mu R_\mu(T;\xi).
\end{equation}
The $\Lambda$ amplitudes are chosen such that
\begin{equation}
    \frac{\partial \mathcal L}{\partial t_\mu}=0
    \quad \text{for all } \mu .
\end{equation}
Therefore, when the total derivative is taken, the terms proportional to
$
\partial t_\mu/\partial \xi
$
cancel,
\begin{equation}
    \frac{dE}{d\xi}
    =
    \frac{d\mathcal L}{d\xi}
    =
    \frac{\partial \mathcal L}{\partial \xi}
    +
    \sum_\mu
    \frac{\partial \mathcal L}{\partial t_\mu}
    \frac{\partial t_\mu}{\partial \xi}
    +
    \sum_\mu
    \frac{\partial \mathcal L}{\partial \lambda_\mu}
    \frac{\partial \lambda_\mu}{\partial \xi}
    =
    \frac{\partial \mathcal L}{\partial \xi}.
\end{equation}
In this sense, $\Lambda$ absorbs the implicit response of the right amplitudes, and converts the derivative of a non-variational CC energy into the partial derivative of a stationary Lagrangian.

\subsection{Expectation-value CCSD and one-particle properties\label{si:xccsd}}
While the CCSD energy depends only on the excitation amplitudes $t_i^a$ and $t_{ij}^{ab}$, general molecular observables require expectation values of operators. In particular, one-particle properties such as the dipole moment and the electron density are determined by the one-particle reduced density matrix (1-RDM). In this work, we use expectation-value coupled cluster with singles and doubles (XCCSD) \citep{korona_one-electron_2006} evaluated from the predicted $T$ amplitudes as a natural baseline for such properties. This provides a simple route from $T_1$ and $T_2$ to approximate correlated observables, against which we compare the more accurate properties obtained when both $T$- and $\Lambda$-amplitudes are available.

In XCC, the normalized expectation value of a one-body operator $\hat{X}$ is written as
\begin{equation}
    \langle \hat{X} \rangle
    =
    \frac{\bra{\Phi_{\mathrm{HF}}} e^{\hat{T}^\dagger} \hat{X} e^{\hat{T}} \ket{\Phi_{\mathrm{HF}}}}
    {\bra{\Phi_{\mathrm{HF}}} e^{\hat{T}^\dagger} e^{\hat{T}} \ket{\Phi_{\mathrm{HF}}}},
\end{equation}
with $\hat{T}=\hat{T}_1+\hat{T}_2$ in CCSD. This expression can be recast into a connected form through an auxiliary operator $\hat{S}$ constructed from $\hat{T}$ by a commutator expansion. In practice, this expansion is truncated, yielding an approximate but size-extensive expectation-value formalism. Here, we use this low-order XCCSD treatment as a baseline method for extracting one-particle properties from predicted $T$ amplitudes alone.

The most general one-particle operator has the second-quantized form
\begin{equation}
    \hat{X} = \sum_{pq} X_{pq}\, \hat{a}_p^\dagger \hat{a}_q,
\end{equation}
where $X_{pq}$ are matrix elements in the molecular-orbital basis. Its expectation value is fully determined by the 1-RDM,
\begin{equation}
    \gamma_{pq} = \langle \hat{a}_p^\dagger \hat{a}_q \rangle,
\end{equation}
since
\begin{equation}
    \langle \hat{X} \rangle = \sum_{pq} X_{pq}\, \gamma_{pq}.
\end{equation}
Thus, once an approximate XCCSD 1-RDM has been constructed from the amplitudes, any one-particle observable follows by contraction with the corresponding operator matrix.

Two important examples are the dipole moment and the real-space electron density. The electronic contribution to the dipole moment is obtained from the dipole-integral matrix $\mu_{pq}^{(\alpha)}$ for Cartesian component $\alpha \in \{x,y,z\}$ as
\begin{equation}
    \mu_\alpha^{\mathrm{el}}
    =
    - \sum_{pq} \mu_{pq}^{(\alpha)} \gamma_{pq},
\end{equation}
to which the nuclear contribution is added in the usual way,
\begin{equation}
    \mu_\alpha
    =
    \sum_A Z_A R_{A\alpha} + \mu_\alpha^{\mathrm{el}}.
\end{equation}
Likewise, the electron density is obtained from the diagonal of the real-space 1-RDM,
\begin{equation}
    \rho(\mathbf{r})
    =
    \sum_{pq} \gamma_{pq}\, \psi_p(\mathbf{r}) \psi_q(\mathbf{r}).
\end{equation}
Therefore, predicted $T$ amplitudes alone already provide access to approximate dipoles, densities, and other one-particle observables through XCCSD.

Because XCCSD relies only on the right-hand amplitudes and a truncated expectation-value construction, its accuracy is limited relative to formulations that also include the left-hand coupled-cluster information. This motivates our extension to models that predict the $\Lambda$ amplitudes in addition to $T$, enabling substantially more accurate one-particle properties.

\subsection{Polarizability\label{si:polarizability}}

We compute the static frozen-orbital polarizability analytically from the coupled-cluster response state. The implementation follows Psi4 \citep{smith2020psi4}, but is implemented in our PySCF-compatible GPU codebase. In this setting, the Hartree--Fock orbitals are held fixed under the external electric field, and the polarizability measures the correlated response of the CCSD amplitudes and $\Lambda$-state density matrices in that fixed molecular-orbital basis.
Following the Psi4 documentation recommendation, we use this frozen-orbital convention for the coupled-cluster response comparison reported here.

Let $\mathbf F=(F_x,F_y,F_z)$ denote a homogeneous electric field. The field-dependent Hamiltonian is
\begin{equation}
    \hat H(\mathbf F)
    =
    \hat H(0)
    -
    \sum_{\alpha\in\{x,y,z\}}
    F_\alpha \hat \mu_\alpha ,
\end{equation}
where $\hat \mu_\alpha$ is the $\alpha$ component of the total dipole operator. The frozen-orbital polarizability tensor is
\begin{equation}
    \alpha_{\alpha\beta}^{\mathrm{FO}}
    =
    -
    \frac{\partial^2 E(\mathbf F)}
    {\partial F_\alpha \partial F_\beta}
    \bigg|_{\mathbf F=0}
    =
    \frac{\partial \mu_\alpha}
    {\partial F_\beta}
    \bigg|_{\mathbf F=0}.
\end{equation}
Because the orbitals are fixed, no coupled-perturbed Hartree--Fock orbital-response terms are included.

For CCSD, the derivative is evaluated through the coupled-cluster Lagrangian,
\begin{equation}
    \mathcal L(T,\Lambda;\mathbf F)
    =
    \bra{\Phi_0}
    (1+\hat\Lambda)
    e^{-\hat T}
    \hat H(\mathbf F)
    e^{\hat T}
    \ket{\Phi_0}.
\end{equation}
At stationarity with respect to $T$ and $\Lambda$, field derivatives of the energy can be written as partial derivatives of the Lagrangian. The first derivative gives the relaxed dipole moment,
\begin{equation}
    \mu_\alpha
    =
    -
    \frac{\partial \mathcal L}
    {\partial F_\alpha}
    =
    \sum_A Z_A R_{A\alpha}
    -
    \sum_{pq}
    \mu^{(\alpha)}_{pq}\gamma_{pq},
\end{equation}
where $\gamma$ is the fixed-orbital CCSD one-particle density matrix. The polarizability is then obtained from the analytic derivative of this density with respect to the perturbing field,
\begin{equation}
    \alpha_{\alpha\beta}^{\mathrm{FO}}
    =
    -
    \sum_{pq}
    \mu^{(\alpha)}_{pq}
    \frac{\partial \gamma_{pq}}
    {\partial F_\beta}.
\end{equation}
The required density response is computed by solving the fixed-orbital CCSD response equations for each Cartesian perturbation direction $\beta$ and contracting the resulting response intermediates with the dipole-integral matrices. 
For the HF and MP2 baselines, we follow Psi4 and compute polarizabilities by finite differences rather than by an analytic response implementation. 
In the main text we report the Frobenius error
\begin{equation}
|\alpha_{\text{pred}} - \alpha_{\text{CCSD}}|_F 
=
\sqrt{\sum_{i,j} (\alpha_{ij}^{\text{pred}} - \alpha_{ij}^{\text{CCSD}})^2}
\end{equation}


\section{Experimental details}

\subsection{Hyperparameters for joint $T$- and $\Lambda$-amplitude training}
\label{si:hyperparameters_neurips_lambda}

We summarize the hyperparameters for \molelambda{} in tab.~\ref{si:tab:hyperparameter_neurips_lambda}. Compared to \citep{thiede2026mole}, we increase the equivariant transformer width and readout capacity and train jointly on the $\Lambda_1$ and $\Lambda_2$ amplitude targets in addition to the $T$-amplitude readout configuration. We opt for a wider latent representation, using $256$ channels for each of the $\ell=0,1,2$ hidden irreps, a larger attention bottleneck, and higher-capacity single-to-pair readouts. The pair-to-quadruple stage is changed from an MLP readout to a dot-product readout, which directly contracts pair features over atoms and feature channels to produce scalar quadruple amplitudes. 
We switch from post-norm to pre-norm transformer blocks, applying equivariant normalization before the attention and message-passing updates to improve training stability. We increased the radial basis from 10 to 16 Bessel functions and the cutoff from 4.0 to 6.0, as we found both improve generalization to out-of-equilibrium geometries.
We also update optimization to use Muon with mixed bfloat16 precision and cosine learning-rate decay with warmup. Our model is trained for up to a week on a single H100.

\begin{table}[t]
\centering
\small
\setlength{\tabcolsep}{4pt}
\renewcommand{\arraystretch}{1.1}
\begin{tabular}{ll}
\hline
\textbf{Category} & \textbf{Configuration} \\
\hline
\multicolumn{2}{l}{\textbf{Transformer}} \\
Number of transformer layers & 4 \\
Hidden irreps & \texttt{256x0e + 256x1o + 256x2e} \\
Edge irreps & \texttt{1x0e + 1x1o + 1x2e} \\
Normalization & Separable equivariant layer norm \\
\multicolumn{2}{l}{\textbf{Radial / GNN block}} \\
Maximum radius & 6.0 \\
Radial basis & Bessel \\
Number of Bessel functions & 16 \\
Polynomial cutoff order & 5 \\
MACE correlation order & 3 \\
\multicolumn{2}{l}{\textbf{Attention block}} \\
Attention type & Sign-equivariant dot-product attention \\
Latent irreps & \texttt{64x0e + 64x1o + 64x2e} \\
Number of attention heads & 8 \\
Norm & Pre-norm with skip connections \\
\multicolumn{2}{l}{\textbf{$T_1$ and $\Lambda_1$ readouts}} \\
Readout module & MLP single-to-pair readout \\
Readout irreps & \texttt{64x0e + 64x1o + 64x2e} \\
MLP hidden width & 64 \\
\multicolumn{2}{l}{\textbf{$T_2$ and $\Lambda_2$ readouts}} \\
Single$\rightarrow$pair module & Tensor-product pair feature module \\
Single$\rightarrow$pair irreps & \texttt{64x0e + 64x0o + 64x1e + 64x1o + 64x2e + 64x2o} \\
Pre-tensor-product irreps & \texttt{64x0e + 64x1o + 32x2e} \\
Pair$\rightarrow$quadruple module & Dot-product readout \\
\multicolumn{2}{l}{\textbf{Training}} \\
Epochs & 100 \\
Batch size & 1 \\
Optimizer & Muon \\
Base learning rate & $10^{-2}$ \\
Muon learning rate & $10^{-2}$ \\
Weight decay & 0 \\
Loss function & summed MSE over amplitudes \\
$\Lambda_1$ loss factor & 1.0 \\
$\Lambda_2$ loss factor & 1.0 \\
Precision & bfloat16 mixed precision \\
\multicolumn{2}{l}{\textbf{Learning-rate schedule}} \\
Scheduler & Cosine annealing \\
Minimum learning rate & $10^{-6}$ \\
Warmup fraction & 0.1 \\
Warmup start factor & 0.01 \\
\hline
\end{tabular}
\normalsize
\vspace*{0.1cm}
\caption{Model architecture and training hyperparameters for the joint $T$- and $\Lambda$-amplitude model.}
\label{si:tab:hyperparameter_neurips_lambda}
\end{table}

\paragraph{$L_2$ and $T_2$ dot-product readout}
Compared to \mole{} \citep{thiede2026mole}, the pair-to-quadruple stage is changed from an MLP readout to a simpler dot-product readout.
Both variants take pair features
$f_{ia}, f_{jb} \in \mathbb{R}^{B \times N_o \times N_v \times N_{\mathrm{atom}} \times F}$
and return scalar quadruple amplitudes
$p_{ijab} \in \mathbb{R}^{B \times N_o \times N_o \times N_v \times N_v}$.
The dot-product readout computes this scalar directly as
\[
p_{ijab}
= \frac{1}{\sqrt{F}}
\sum_{\alpha=1}^{N_{\mathrm{atom}}}
\sum_{c=1}^{F}
f_{ia\alpha c} f_{jb\alpha c},
\]
i.e. it contracts over the atom axis and the full pair-feature channel axis. In the MLP variant, the pair features are first normalized over the combined $(N_{\mathrm{atom}},F)$ axis and projected into a hidden equivariant representation, e.g.
$32 \times 0e + 16 \times 1e + 8 \times 2e$ with total feature dimension
$32 + 16 \cdot 3 + 8 \cdot 5 = 120$. An invariant contraction is then performed blockwise over atoms and spherical components, producing one scalar per multiplicity channel,
$32 + 16 + 8 = 56$ invariant features per $(i,j,a,b)$ quadruple. These 56 features are passed through a small MLP, for example $56 \rightarrow 16 \rightarrow 1$, to produce the final scalar amplitude.
Empirically we find that the dot-product readout reaches almost the same accuracy, at $~30\%$ faster training and inference.

\subsection{MLIP baselines\label{si:mlip}}
For energy and force prediction we compare against state-of-the-art machine learning interatomic potentials (MLIPs) that are trained on the same five QM7 training-set sizes (100, 300, 1000, 3000, and full 5732) for both direct CCSD targets and residual targets $\Delta=\mathrm{CCSD}-\mathrm{MP2}$. For the CCSD target prediction we use fixed atomic reference energies; for the delta target we set the atomic references to zero. All baseline runs are trained on a single NVIDIA A100 80GB GPU for 3000 epochs and follow the standard train/validation splits used throughout our QM7 experiments.

\textbf{MACE:}
For MACE, we use three interaction blocks, hidden irreps \texttt{128x0e + 128x1o}, cutoff radius $r_{\max}=6.0$~\AA, and spherical harmonics up to $\ell=3$. Optimization uses AdamW with batch size 32, learning rate $10^{-3}$, weight decay $10^{-3}$, cosine annealing to $10^{-6}$, and 3000 epochs. The loss weights are $(1,100,1)$ for energy, forces, and dipoles.  Across the 10 runs, the total training time is approximately 150 A100-hours.

\textbf{eSEN:}
For eSEN, we use the UMA \texttt{K4L2} backbone with 128 sphere channels, $l_{\max}=2$, $m_{\max}=2$, cutoff radius 6~\AA, and at most 30 neighbors. Training uses AdamW with learning rate $5\times10^{-4}$, gradient clipping of 10, zero weight decay, bf16 precision, and max-atoms batching. We supervise direct energies and forces with coefficient 10 for both terms. The ten runs account for roughly 110 A100-hours.


\subsection{Evaluation set selection\label{si:eval_set_selection}}
\paragraph{PubChem.}
The PubChem size-extrapolation set was drawn from the PubChem Compound DrugBank \citep{kim2025pubchem,wishart2018drugbank}.
We downloaded the DrugBank information-source subset in JSON format with coordinate type set to 3D, then kept even-electron molecules containing only H, C, N, and O, with available conformer coordinates and exactly 14 heavy atoms.
The random selection used NumPy seed 42.
The final 100 CIDs are
62787, 46937113, 5289331, 604519, 5192, 2482, 11229234, 680935, 5288483, 1318,
54695768, 9859093, 7184, 6199, 26757, 445245, 122282, 25429, 736186, 4099,
62529, 161358, 6852125, 6862, 445210, 450268, 2148, 697959, 6047, 3052,
7045767, 3604, 5280980, 123831, 3845, 441350, 1616, 21120522, 12551, 3870203,
30487, 445211, 445994, 6420149, 45098889, 13294447, 25145077, 3518, 6839, 33037,
91514, 15664, 9815560, 61810, 16615, 445858, 4284, 6950274, 3102, 90274764,
5494443, 254, 7127816, 3744, 1082702, 38945, 2519, 2737071, 12192007, 198695,
5288636, 5447130, 446429, 441314, 121026, 1615, 3698, 1676, 4474777, 60748,
445825, 44585557, 16143, 104904, 5494411, 11071, 3893, 40632, 8367, 46936338,
10130337, 12097317, 10205, 17754076, 447364, 25058136, 61361, 5394, 14052, and 178052.

\paragraph{Amino acids.}
The amino-acid set contains the following 18 standard amino acids:
alanine, arginine, asparagine, aspartate, glutamine, glutamate, glycine, histidine, isoleucine, leucine, lysine, phenylalanine, proline, serine, threonine, tryptophan, tyrosine, and valine.

\section{Additional Results}

\paragraph{On-top pair-density}
\begin{figure}
    \centering
    \includegraphics[width=0.69\linewidth]{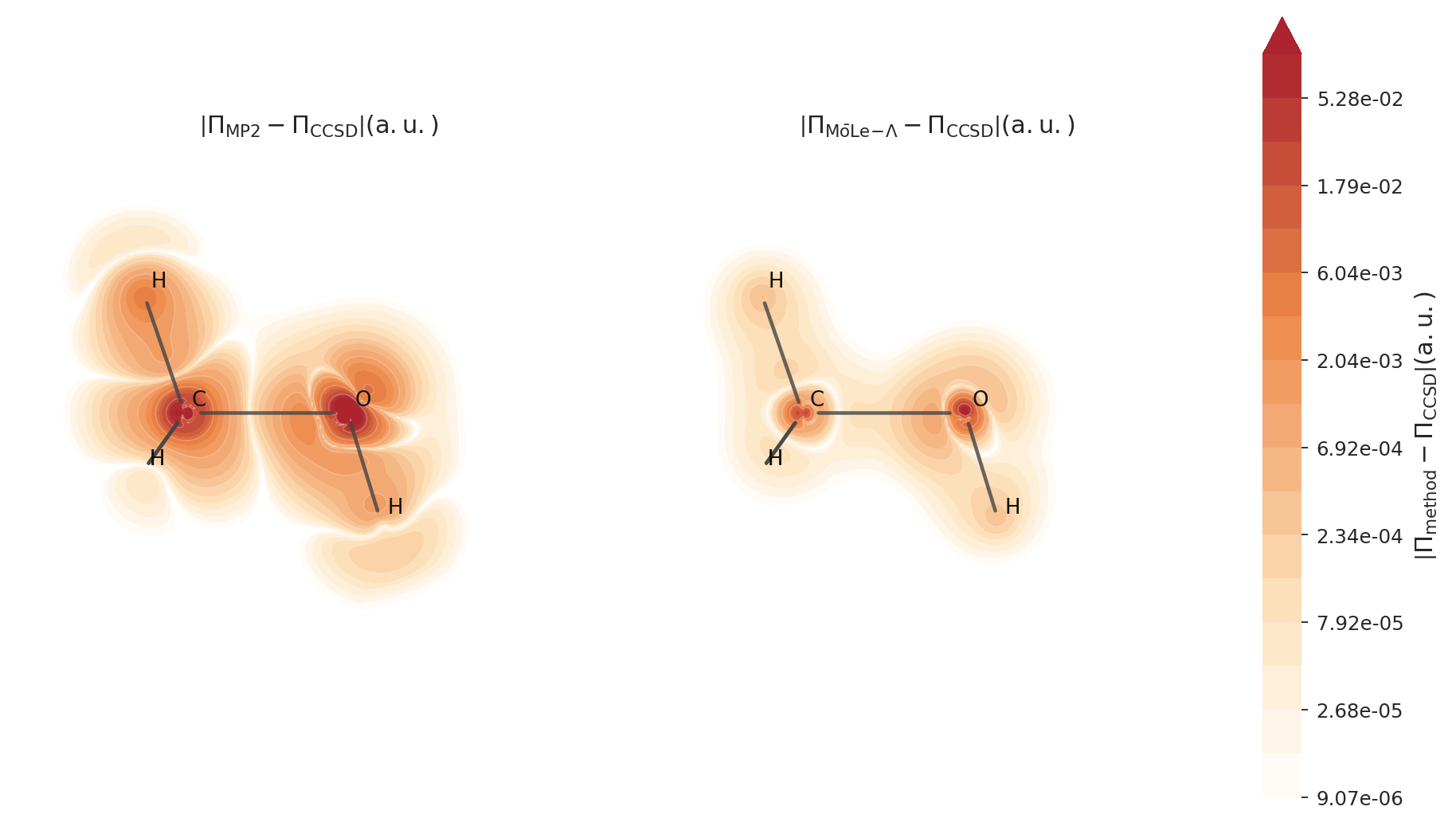}
    \caption{
    Error of the on-top pair-density $\Pi$, obtained from the 2-RDM, of MP2 and \molelambda{} compared to CCSD.
    }
    \label{fig:ontoppairdensity}
\end{figure}
Fig.~\ref{fig:ontoppairdensity} shows the on-top pair-density error, $\left|\Pi_{\mathrm{method}}(\mathbf r,\mathbf r)-\Pi_{\mathrm{CCSD}}(\mathbf r,\mathbf r)\right|$, for MP2 and \molelambda{} in methanol. The MP2 error is broad and pronounced, with extended high-error regions around both heavy atoms and along the C--O bond, indicating a substantial misrepresentation of short-range electron correlation across the molecular framework. By contrast, for \molelambda{} the largest errors remain localized near the atomic centres, while the diffuse bond-centred and off-atom error visible for MP2 is strongly suppressed. This indicates that \molelambda{} reproduces the CCSD on-top pair density much more faithfully, reducing the overall magnitude of the error and recovering its correct spatial localization.



\end{document}